# Left Ventricle Segmentation in Cardiac MR Images Using Fully Convolutional Network

Mina Nasr-Esfahani, Majid Mohrekesh, Mojtaba Akbari, S.M.Reza Soroushmehr,
Ebrahim Nasr-Esfahani, Nader Karimi, Shadrokh Samavi, Kayvan Najarian

*Abstract*— **Medical image analysis, especially segmenting a specific organ, has an important role in developing clinical decision support systems. In cardiac magnetic resonance (MR) imaging, segmenting the left and right ventricles helps physicians diagnose different heart abnormalities. There are challenges for this task, including the intensity and shape similarity between left ventricle and other organs, inaccurate boundaries and presence of noise in most of the images. In this paper we propose an automated method for segmenting the left ventricle in cardiac MR images. We first automatically extract the region of interest, and then employ it as an input of a fully convolutional network. We train the network accurately despite the small number of left ventricle pixels in comparison with the whole image. Thresholding on the output map of the fully convolutional network and selection of regions based on their roundness are performed in our proposed post-processing phase. The Dice score of our method reaches 87.24% by applying this algorithm on the York dataset of heart images.**

## I. Introduction

Medical image analysis and segmentation are important challenges in the image processing field that requires highly accurate results as it deals with human health. A typical example is the segmentation of left ventricle (LV) in cardiac magnetic resonance imaging (CMRI). Some cardiac properties such as ejection fraction, LV volume and wall thickness help physicians evaluate the cardiovascular function and diagnose the disease. As manual segmenting is a tedious, error prone and time consuming task, automatic or semi-automatic segmentation would help physicians in their decision making. However, there are some challenges facing the segmentation of LV in CMRI. One of the basic challenges in segmentation of LV in CMRI is the non-negligible overlapping of cardiac objects intensity distribution and that of surrounding background. Shape of LV varies across slices and phases despite the extreme little number of pixels belonging to the heart in comparison with other pixels of the frame. The boundaries are very inaccurate specially in presence of inherent noise in real-time imaging of the heart called cine MRI[1]. A number of methods have been proposed in the last decade to automatically calculate the mentioned LV properties. These methods can be categorized to three groups: 1) The methods that use prior information 2) The methods based on automatic localization of the heart and 3) The methods that employ segmentation techniques. In the category of segmentation there are many techniques such as region growing, edge detection, deformable model, active shape model (ASM), active appearance model (AAM), atlas based model and pixel classification model [2-8]. In [3], a region growing method is used for LV segmentation which is initialized by several seed points and grows the region according to some features. Deformable models such as active contour are initialized by a predefined contour and minimize an energy function to fit the contour to the border of object [4]. ASM [5,6] and AAM [7] use a dataset with manually segmented LV as a training set. The result of the training phase is a model which represents LV and try to fit this model to the target image in a test phase. Other methods in this category are atlas-based methods [8] that extract prior information from a reference image called atlas. In medical image processing, an atlas gives an estimation of object positions, shape, texture, and the features of adjacent objects in images [9]. Pixel classification methods such as clustering and neural network (NN), classify the pixels with similar features into one group. New generation of deep NN called convolutional NN (CNN) which is designed for image analysis has been widely used for image segmentation. The method proposed in [10] use eight convolutional layers for extracting the region of interest (ROI), then segmentation of right ventricle is performed by another five convolutional layers of network. CNNs are supervised training models which are trained to learn hierarchies of features automatically that yield performance in robust classification [11]. Recent successes of CNNs in different image segmentation applications prove their abilities to solve different image segmentation problems. New architectures of CNNs such as Alex-Net [12], VGG-Net [13], Google-Net [14] and Res-net [15], have made the CNN the de facto standard for classification in many applications [1]. Another CNN designed especially for biomedical image segmentation is U-Net. Architecture of this network is similar to a U-shape and consists of a contracting path to capture context and a symmetric expanding path that enables precise localization of biomedical objects [16].

In this paper we propose an LV segmentation method. The proposed method has four main stages: 1) Pre-processing 2) ROI extraction that is based on heart movement, 3) Training a fully convolutional network (FCN) [11] which is a new architecture of CNN designed for semantic segmentation

M. Nasr-Esfahani, M. Mohrekesh, M. Akbari and N. Karimi are with the Department of Electrical and Computer Engineering, Isfahan University of Technology, Isfahan 84156-83111, Iran.

S.M. R. Soroushmehr is with the Department of Computational Medicine and Bioinformatics and the Michigan Center for Integrative Research in Critical Care (MCIRCC), University of Michigan, Ann Arbor, MI, U.S.A.

S. Samavi is with the Department of Electrical and Computer Engineering, Isfahan University of Technology, Isfahan 84156-83111, Iran. He is also with the Department of Emergency Medicine, University of Michigan, Ann Arbor, MI, U.S.A.

K. Najarian is with the Department of Computational Medicine and Bioinformatics, the Michigan Center for Integrative Research in Critical Care (MCIRCC) and the Emergency Medicine Department, University of Michigan, Ann Arbor, MI, U.S.A.

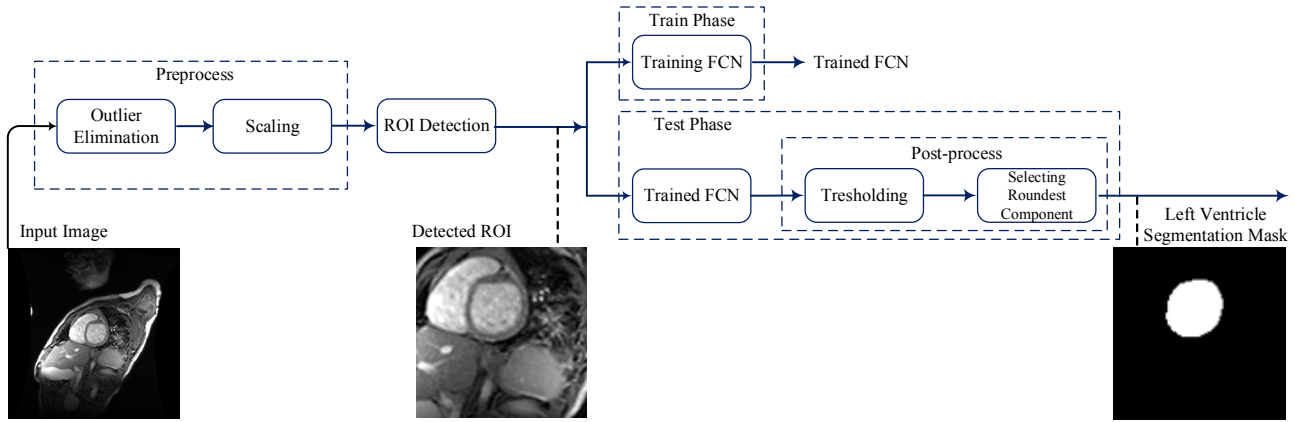

Figure 1: Overall block diagram of the proposed algorithm.

and 4) Post-processing which selects a proper region based on its roundness. Details of the proposed method are explained in section II. The evaluation of the proposed LV segmentation method is presented in section III where we use three metrics, accuracy, Dice and sensitivity for this purpose.

## II. PROPOSED METHOD

The block diagram of the proposed method is shown in Fig. 1. This method contains four steps:
1. The preprocessing which statistically normalizes and adjusts images.
2. ROI extraction which is done based on heart movement in a cardiac cycle.
3. The training of the FCN by the ROI, and using it in the test phase.
4. Post-processing for selecting the proper component as the LV.

### A. Preprocessing

Here we use images of a database available in York university [17]. These images have different range of intensity (0 to 8011) and are affected with the inherent noise of cine MRI. Therefore, before applying each segmentation algorithm, they should be in the same range of intensities. For this purpose, we preprocess the images in two steps including outlier elimination and scaling.

Noise of cine MRI is dominant to the intensities of pixels and this fact is obvious in histogram of the frames. This fact causes very high values of intensity in diagrams of histogram. Hence, we first eliminate the one percent of outlier intensities and replace their intensities with the nearest value not present in the outlier set of intensity values. After this process, intensity scale is still different due to the intensity variety before normalizing. Thus, we then scale the resulting data into new [0, 1] range to be appropriate for the input layer of the FCN network

### B. Automatic ROI extraction

Each sequence in our database contains 20 frames for a cardiac cycle. Heart movement is our key point for ROI extraction in each sequence, but it's movement is very weak in transition between two consecutive frames which makes ROI detection so challenging. Block diagram of finding ROI is shown in Fig. 2. Our proposed ROI detection method is described in five steps as follows:

1. *Absolute difference of each two consecutive frames*: The result specifies the variation of intensities between two consecutive frames, which includes heart movement along with a lot of noise.

2. *Sum of all absolute differences*: Due to the high amount of noise in each image of absolute difference, heart is not very explicating part in resulting absolute difference. Thus, aggregating all of absolute differences related to a sequence causes heart become more salient in the resulting sum.

3. *Saliency detection*: Based on saliency algorithm which is proposed in [18] the saliency map of previous step is calculated.

4. *Thresholding:* Thresholding on saliency map and extract the 90% of more salient parts is performed.

5. *Applying bounding box:* A square is determined as ROI by applying a bounding box around the salient part.

### C. Fully Convolutional Network

FCN [15] is a generation of CNN which produces hierarchies of features. Basic components of FCN are convolution, pooling and deconvolution layers. FCN input is an image with an arbitrary size and data in each FCN layer is a matrix with the size of $h \times w \times d$ where $h$ and $w$ are the spatial sizes of feature map and $d$ is the number of feature maps in each layer. We can also consider the input as a feature map of input layer of FCN with the pixel size of $h \times w$ and $d$ color channels. The architecture of the FCN consists of a contracting path in convolution and pooling layers and expanding path in deconvolution layers. Deconvolution layers classify extracted feature maps with progressive upsampling until reaching the size of input image. There are three types of FCN called FCN-32s, FCN-16s and FCN-8s different in selection of deconvolution feature maps. It is shown in [11] that FCN-8s is the best choice for semantic segmentation and works better than other versions. Therefore, we use this type of FCN for left ventricle segmentation.

### D. Post-processing

The FCN output is a probability map which specifies the amount of membership to target for each pixel and a thresholding is needed to be applied on FCN map. This threshold is obtained using Otsu algorithm [19] which divides the probability map into two categories with minimum variance in each one. Obtained region using the above thresholding might consist of more than one connected component and hence the target needs more process to be specified. In all of the test images, some parts of the LV or the entire LV might exceed the threshold and hence they have intersection with the resulting region. In order to identify which part is LV, we propose the region roundness parameter for selecting the proper region as LV. Circle has uniform distance from its center and rounder regions have more uniform distance from their center of gravity. Equation 1 shows our proposed roundness function:

$$Roundness = \frac{\sigma_D}{\mu_D} \quad (1)$$

$D$ is the set of distances for one connected component, $\sigma$ is the standard deviation and $\mu$ refers to the average of values. Finally, the region with more roundness property is selected as the LV.

### III. EXPERIMENTAL RESULT

### A. Dataset

For our experiments we use a dataset that is publicly available in the York University website and contains CMRI of 33 patients. Each set has 8 to 15 sequences provided by the department of Diagnostic Imaging of the Hospital for Sick Children in Toronto, Canada [17]. Each sequence is a cardiac cycle with 20 frames with manual segmentation of endocardium for most of the frames. Image resolution is 256×256 pixels and each sequence is captured in one cardiac cycle. Two of the 33 patients have unknown diagnosis and the others are diseased [6].

### B. Evaluation

The quantitative evaluation of the proposed method performance is based on three common metrics which are accuracy, Dice and true positive rate (TPR) / sensitivity defined in (2), (3) and (4) respectively. All metrics are calculated by counting pixels belonging to each category. Table I shows effects of each step in our proposed method for segmentation of LV.

$$Accuracy = \frac{TP + TN}{TP + FP + FN + TN} \quad (2)$$

$$Dice\ Score = \frac{2TP}{2TP + FP + FN} \quad (3)$$

$$TPR = \frac{TP}{TP + FN} \quad (4)$$

In Table II, we compare our proposed method with the methods proposed in [5], [6] and [7] using the same dataset. The method of [5] is the standard ASM and the one proposed in [7] is a modified AAM. Moreover, a version of ASM for segmenting the LV is presented in [6]. The results of [5] and [7] are obtained from [6].

The original images, ROI, FCN output, post-processed output and error of segmentation for some sample frames are shown in Fig 3. The error shows the difference between the result of the proposed method final output and the ground truth which is obtained from physicians' manual segmentation. For the input image shown in the first row of Fig. 3, FCN works properly and the error is very little. The second row has a few organs similar to the LV and FCN segments those ones as LV wrongly. However, the post-processing step corrects them. The third row is an image with a lot of noise and small LV. As can be seen, the proposed method produces acceptable results for this image.

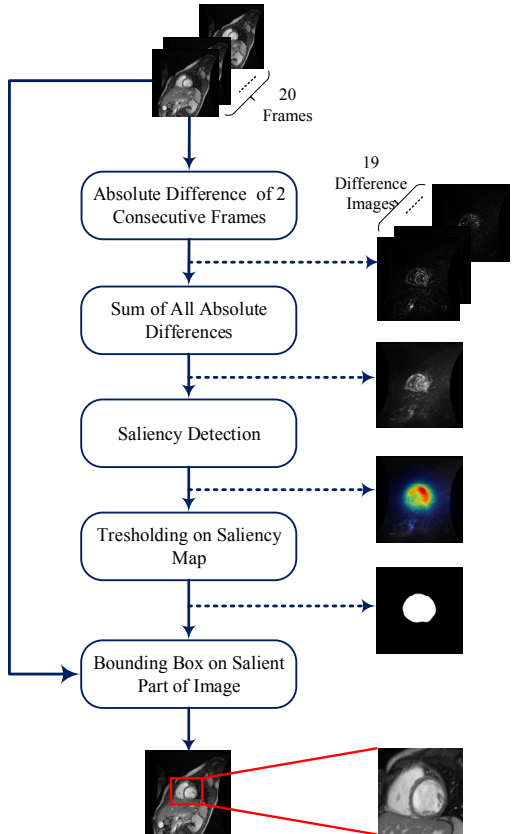

Figure 2. Block diagram of ROI extraction

Table I. Effect of each steps in our proposed method

|  | Accuracy | Dice | Sensitivity |
|---|---|---|---|
| FCN on images | 99.57 | 76.24 | 71.20 |
| FCN on ROI | 98.34 | 86.94 | **87.75** |
| FCN on image + post-process | **99.64** | 79.51 | 71.62 |
| FCN on ROI + post-process | 98.39 | **87.24** | 87.69 |

Table II. Comparison with [5], [6], [7]

| Method | Dice |
|---|---|
| ASM [5] | 73.1 |
| EM-RASM [6] | 79.4 |
| AAM [7] | 81.7 |
| Proposed method | 87.24 |

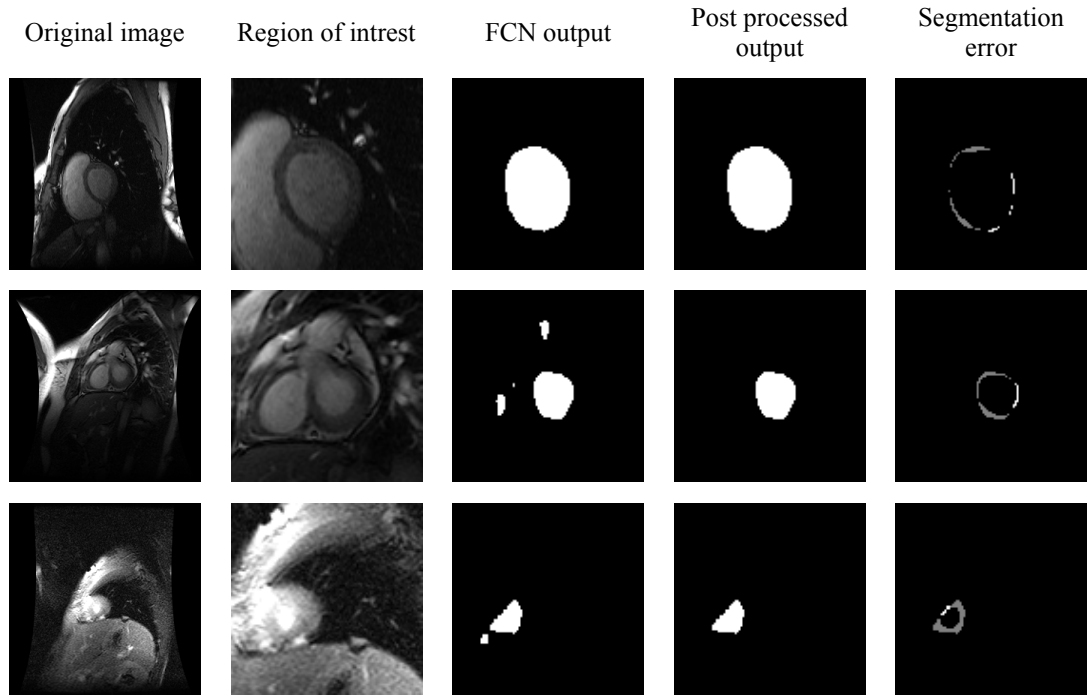

Figure 3. Produced outputs for three sample frames.

IV. CONCLUSION

In this paper we proposed a method for segmentation of left ventricle in CMRI images. Our proposed segmentation method contains three main steps. In the first step we extract the region of interest by using the movement of heart in the frames of a cycle. In the second step we used fully convolutional network for segmenting candidate regions. Third step of our proposed method contains selecting the proper region based on the roundness parameter for candidate. We applied our proposed method on a publicly available database and achieved the Dice score of 87.24.